\documentclass[review]{elsarticle}

\usepackage{lineno,hyperref}
\modulolinenumbers[5]
\usepackage{multirow}
\usepackage{booktabs}
\usepackage{diagbox}
\usepackage{subfig}
\usepackage{soul}
\setstcolor{orange}
\usepackage{color}
\usepackage{hyperref}
\usepackage[table,dvipsnames]{xcolor}
\definecolor{ao(english)}{rgb}{0.0, 0.5, 0.0}
\definecolor{brown(web)}{rgb}{0.65, 0.16, 0.16}
\journal{Journal of \LaTeX\ Templates}

\begin{document}

\begin{frontmatter}

\title{Enhancing Egocentric 3D Pose Estimation \\ with Third Person Views}

\author{Ameya Dhamanaskar, Mariella Dimiccoli, Enric Corona, Albert Pumarola, Francesc Moreno-Noguer}
\address{Institut de Robòtica i Informàtica Industrial, CSIC-UPC \\ Carrer Llorens i Artigas 4-6, 08028, Barcelona, Spain}

\begin{abstract}
In this paper, we propose a novel approach to enhance the 3D body pose estimation of a person computed from videos captured from a single wearable camera. 
The key idea is to leverage high-level features linking first- and third-views in a joint embedding space. To learn such embedding space we introduce \textit{First2Third-Pose}, a new paired synchronized dataset of nearly 2,000 videos depicting human activities captured from both first- and third-view perspectives. We explicitly consider spatial- and motion-domain features,  combined using a semi-Siamese architecture trained in a self-supervised fashion.
Experimental results demonstrate that the joint multi-view embedded space learned with our dataset is useful to extract discriminatory features from arbitrary single-view egocentric videos, without needing domain adaptation nor knowledge of camera parameters. We achieve significant improvement of egocentric 3D body pose estimation performance on two unconstrained datasets, over three supervised state-of-the-art approaches.
Our dataset and code will be available for research purposes \footnote{\url{https://github.com/nudlesoup/First2Third-Pose}}.

\end{abstract}


\end{frontmatter}


\section{Introduction}
\label{sec:intro}
Egocentric vision is raising increasing attention in recent years, since it may enable several important applications in the fields of healthcare, robotics or augmented reality \cite{kutbi2020usability,dimiccoli2018computer,liang2015ar}. For many of these applications, estimating the 3D full body pose of the person wearing the camera is of special interest since it conveys rich information about his/her activities, interactions and behaviour. However, inferring the first-person's 3D body pose from a given egocentric video sequence is a challenging problem in computer vision since wearable cameras are typically worn on the chest or on the head, and have almost no view of the camera wearer’s body. As a consequence, state-of-the-art  approaches for third-person body pose estimation are not suited to the egocentric domain, and dedicated methods need to be developed. 

Despite the relevance of the problem, 
egocentric body pose estimation has received little attention in the literature so far \cite{jiang2017seeing,yuan20183d,yuan2019ego,ng2020you2me,jiang2021egocentric,wang2021estimating}. 
Prior work has shown the importance of leveraging egomotion and the coarse scene structure to predict the body pose behind the camera \cite{jiang2017seeing}. More recently, second person body poses, as observed in a first-person video stream, have been used to enhance egocentric pose estimation during dyadic interactions \cite{ng2020you2me}. While these methods rely only on the egocentric video data itself to estimate the egopose, another line of work uses simulated data to learn a control policy that is ultimately transferred to egocentric videos.
\cite{yuan20183d} used simulated data to learn a control policy accounting for the physics that underlies the kinematics of the motion and hence able to generate physically plausible first-person body poses. In the same spirit, \cite{yuan2019ego} learns a control policy from unsegmented MoCap data without the need of domain adaptation to transfer them to real egocentric data.
Lately, fostered by potential applications in the field of augmented and virtual reality, egocentric 3D pose has been estimated from wearable cameras with eye-fish lens, which ensures a larger body part view \cite{jiang2021egocentric,wang2021estimating} by incorporating motion priors learned from mocap data. 
All these methods, however, either rely solely on information available in first-person videos or leverage simulated data for egopose from existing mocap data or humanoid simulator.

\begin{figure}[t!]
\includegraphics[width=\columnwidth, trim=0mm 220mm 0mm 10mm, clip]{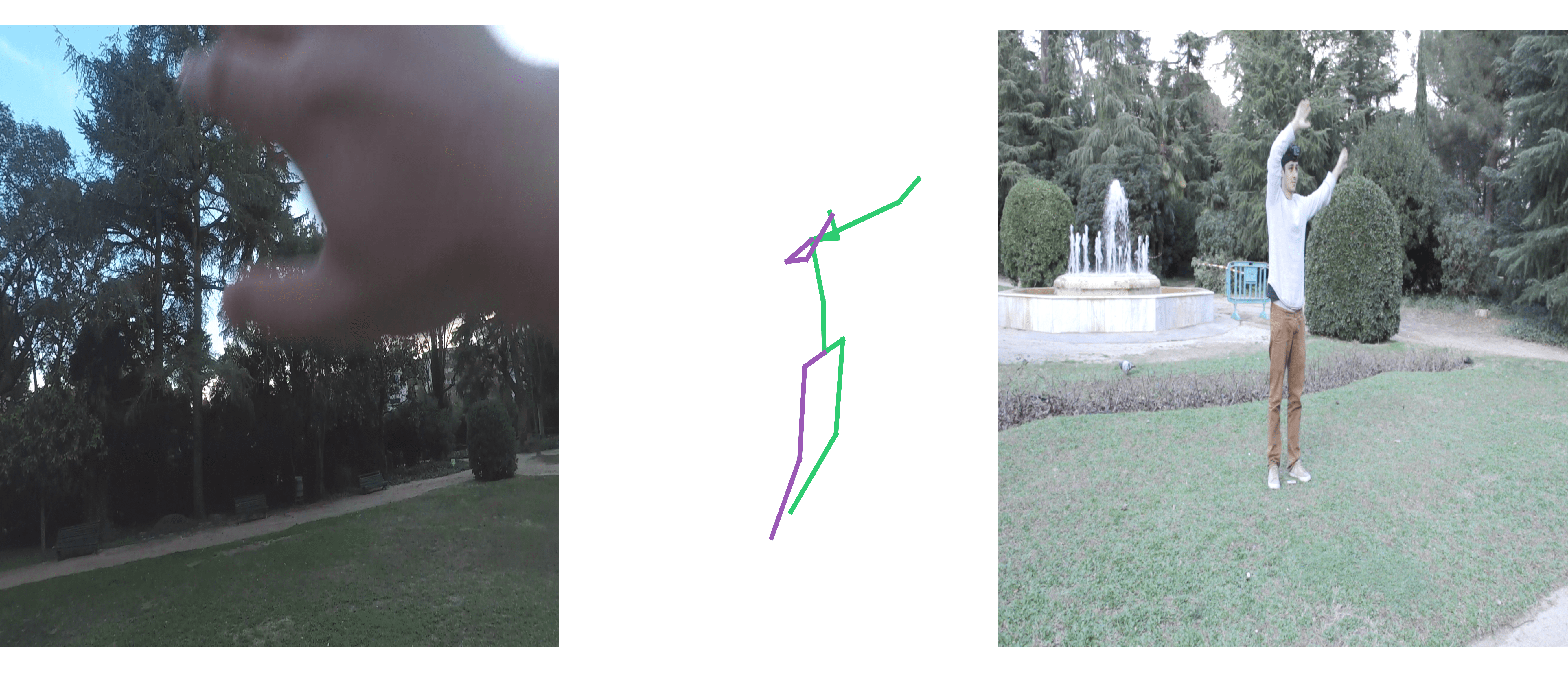}
\vspace{-3mm}
  \caption{First-person (left) and third-person (right) perspectives represent the two sides of a same coin. Our work  considers how a joint embedding space between these two worlds can facilitate egocentric 3D body pose estimation. The skeleton in the center has been estimated from an egocentric video by relying on our joint embedding space.}
  \label{fig:overview}
\end{figure}

In this paper, we explore whether visual cues on third-person videos can help to improve the robustness of 3D pose detectors on first-person views. For this purpose, we rely on a new, 
real dataset of synchronized paired first- and third-person videos to learn a joint embedded space that we leverage to enhance 3D egopose estimation (see Figure \ref{fig:overview}).
To learn the embedding space, we train in a self-supervised fashion a two streams, semi-Siamese convolutional neural network for the task of discriminating if two views (first- and third-views) correspond to the same 3D skeleton. At inference time, the embedded space is used to extract image features allowing to discriminate different 3D human poses from new, unpaired egocentric videos. Our approach does not require domain adaptation to transfer to new unpaired egocentric datasets, and can be applied to egocentric videos captured in the wild, for which the camera parameters are typically unknown.  
 Experimental results on two real datasets show that the use of joint first- and third- embedded features has a significant benefit for 3D egopose estimation. 
Specifically, we reduce the error of 12.71\% and 7.51\% on average respectively on two datasets over three state-of-the-art  approaches.




Our key contributions can be summarized as follows:
\begin{itemize}
\setlength{\itemsep}{0pt}
\item We demonstrate for the first time the benefit of linking first- and third-person views for the task of 3D egocentric body pose estimation.
\item We collect and make publicly available \textit{First2Third-Pose}, a large dataset consisting of nearly 2,000 synchronized first- and third-views videos, capturing 14 people performing 40 different activities in 4 different scenarios. 
\item We show that our dataset is useful to extract discriminative features for estimating 3D egopose from arbitrary egocentric videos, without any knowledge of the camera parameters.
\item We achieve consistent and significant performance improvement on two real datasets,  \textit{First2Third-Pose} and \cite{jiang2017seeing},  over three existing  baseline methods  \cite{jiang2017seeing,ng2020you2me,xiao2018simple}. 
\end{itemize}


\section{Related Work}
\label{sec:related}

\noindent{\bf 3D pose estimation from third-view.}
Learning to estimate 3D body pose from a single RGB image from third-view (assuming the person is seen in the image) is a long-standing problem in computer vision. Most approaches in this area follow a fully-supervised pipeline, and use images annotated with 3D poses to train a deep neural network that regresses the 3D pose directly from images {\cite{li2015maximum,tekin2016structured,dabral2018learning}} or via an intermediate step that estimates the 2D pose {\cite{tome2017lifting,Moreno_cvpr2017,cai2019exploiting}} 
{The architectures explored so far for this task range from Euclidean CNN  to more recent nonEuclidean Graph-Convolutional Network (GCN). Some representative examples are as follows. \cite{li2015maximum} proposed deep neural network approach for maximum-margin structured learning that learns jointly the feature representations for image and pose as well as the score function.
\cite{tome2017lifting} proposes the first
CNN architecture for jointly estimate 2D landmark and 3D human pose.
\cite{cai2019exploiting} models spatial-temporal dependencies between different joints of temporal consecutive skeletons through a GCN approach and consolidate features across scales via a  hierarchical “local-to-global” architecture.}

However, since all these approaches require large amount of annotated data for training, they are typically trained on datasets acquired in controlled indoor environments, for which it is easy to use motion capture systems \cite{mehta2017monocular,ionescu2013human3}. 

Along another body of work devoted to 3D mesh reconstruction, the 3D skeleton is often explicitly taken into account.
Sun et al. \cite{sun2019human} proposed a module for disentangling the skeleton from the rest part of the human 3D mesh, hence building a bridge between 2D/3D pose estimation and 3D mesh recovery. Wang et al. \cite{wang2020sequential} directly infer sequential 3D body models by extracting local features of a sequence of point clouds and regress 3D coordinates of mesh vertices at different resolutions from the latent features of point
clouds. In \cite{rong2020frankmocap} expressive body motion capture including 3d hands, face, and body are estimated from a single image as a form of shape and pose parameters of the SMPL-X 3D model of the human body. \cite{Xu_pami2021} proposed a resolution-aware network for 3D human pose and shape estimation that can handle arbitrary-resolution input with one single model. 

Interestingly, to enable training on 3D body pose datasets acquired in-the-wild, a number of approaches only use 2D weak annotations \cite{sun2017compositional,pavlakos2018ordinal,zhou2016deep} or used weak \cite{iqbal2020weakly,cai2018weakly} and self-supervised based methods \cite{jenni2020self}. 
\cite{iqbal2020weakly}
proposed using unlabeled multi-view data for training in end-to-end manner by enforcing the 3D poses estimated from different views to be consistent. \cite{cai2018weakly} proposed to use depth images captured by commodity RGB-D cameras at training time to alleviate the burden of costly 3D annotations in large-scale real datasets.
\cite{jenni2020self} proposed a self-supervised approach to learn features representations suitable for 3D pose estimation, that uses as pretext task the detection of synchronized views (which are always related by a rigid transformation).

Another self-supervised method for 3D pose estimation was proposed in Rodhin et al. \cite{rhodin2018unsupervised}, where the latent 3D representation is learned by reconstructing one view from another. However, differently from \cite{jenni2020self} their approach strictly relies on the knowledge of the camera extrinsic parameters and background images, and therefore is not suited to datasets captured in the wild.

In this paper, we use first- and third-person paired data not only to get weak annotations for training, but also to learn a multi-view embedding space in a self-supervised fashion, that we further exploit to enhance 3D pose estimations. Differently from \cite{jenni2020self}, where the pretext task translates into a classification of rigid versus non-rigid motion, in our case there is no direct link between the image information of the two types of images (first and third view). In addition, in contrast to \cite{rhodin2018unsupervised}, our approach does not require knowledge of the camera  parameters nor of background images.
%



\vspace{1mm}
\noindent{\bf 3D pose estimation from first-view.}
Inferring human poses from egocentric images or videos is a problem that has been looked into only recently. Early works focused on estimating gestures and hand poses assuming that arms were partially visible \cite{li2013model,li2013pixel,rogez2015first}. In \cite{shiratori2014motion}  several body-mounted cameras  on  person’s joints were used to infer body joint locations via a structure-from-motion approach. \cite{jiang2017seeing} has been pioneering in showing that it is possible to estimate the invisible full body pose of the camera wearer directly from egocentric videos. This work considered dynamic motion signatures and static scene structure cues to build a motion graph from the training data and recovered the pose sequence by solving for the optimal pose path. 
More recently, \cite{ng2020you2me} leveraged the visible body pose of a person interacting with the camera wearer to improve the wearer's pose estimation.

Other methods use a humanoid simulator in a
control-based approach \cite{yuan20183d,yuan2019ego} to estimate the 3D body pose of a camera wearer. \cite{yuan20183d} learns a control policy on simulated data in a two-stage imitation learning process that yields physically valid 3D pose sequences. This is evaluated quantitatively only on synthetic sequences.
On the same line, \cite{yuan2019ego} proposed an approach that can learn a Proportional Derivative control-based policy and a reward function from unsegmented MoCap data and estimate various complex human motions in real-time without the need to perform domain adaptation.


More recent approaches estimate egopose from video captured by a head-mounted and front facing fisheye camera
\cite{jiang2021egocentric,wang2021estimating,xu2019mo,tome2019xr}, which better simulates augmented and virtual devices. Mo2Cap2 \cite{xu2019mo} and xR-egopose \cite{tome2019xr} estimate the local 3D body pose in egocentric camera space, whereas \cite{wang2021estimating} proposes a method to estimate it in the world coordinate system. 
This is achieved by leveraging the 2D and 3D
keypoints from CNN detection as well as VAE-based motion priors learned from a large mocap dataset.
\cite{jiang2021egocentric} leverages on both the dynamic motion information obtained from camera SLAM, and the occasionally visible body parts to predict jointly head and body pose.  
Unlike any of the existing methods, our approach exploits the underlying connection between first- and third-views for 3D egopose estimation.

\vspace{1mm}
\noindent{\bf Linking first-person and  third-person perspectives.}
Previous work has demonstrated the benefits of linking  first-person and third-person perspectives for different tasks.
\cite{soran2014action} showed the potential of combining a single wearable camera and multiple static cameras to better understand action recognition.
More recently, \cite{Sigurdsson_2018_CVPR} introduced a large-scale dataset of paired first- and third-person
videos and used it to learn a joint multi-view representation and transfer knowledge from
the third-person to the first-person domain for the task of zero-shot action recognition. A combination of first-person views from two social partners has been explored for recognizing micro-actions and reactions during social interactions
\cite{yonetani2016recognizing} as well as to improve activity recognition of two partners engaged in the same activity \cite{bambach2015viewpoint}. 
In \cite{fan2017identifying},  an embedding space shared by first- and third-person videos is learned with the goal of matching camera wearers between third and first-person. 
Ego-exo \cite{li2021ego} is a framework to create
strong video representations for downstream egocentric understanding tasks by leveraging traditional third-view large scale datasets. 

In any event, and to the best of our knowledge, the potential of linking the first-person and third-person perspective for 3D egocentric body pose estimation we propose in this paper has never been explored so far.



\begin{figure}[t!]
\begin{center}
\includegraphics[width=0.8\columnwidth, trim=10mm 110mm 20mm 1mm, clip]{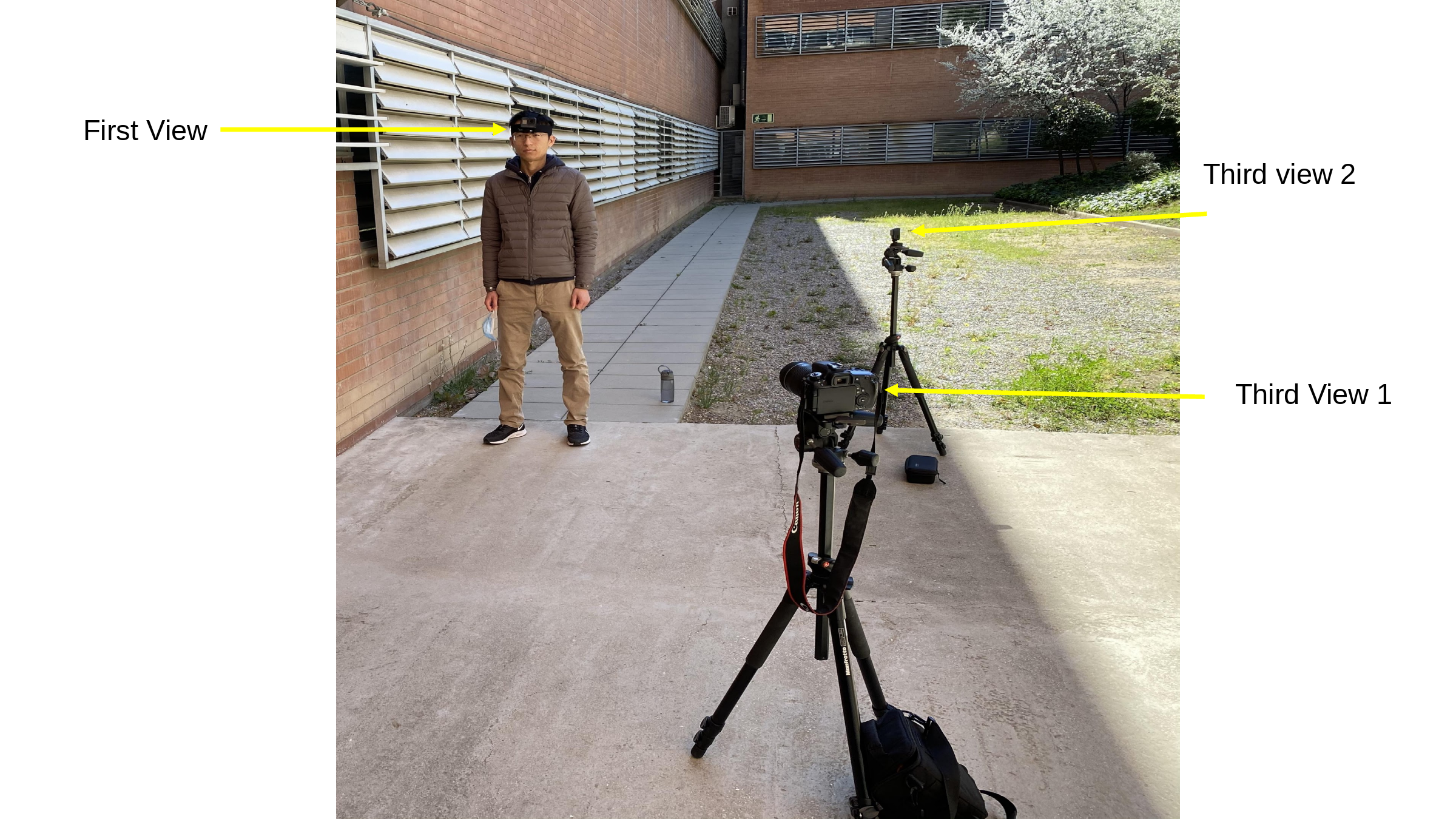}
\end{center}
\vspace{-4mm}
  \caption{Capture setup used for our \textit{First2Third-Pose} dataset. In addition to a head-mounted wearable camera, two static cameras are used to capture side and front views.}
  \label{fig:setup}
\end{figure}

\begin{figure*}[t!]
\begin{center}
\includegraphics[width= 0.9\linewidth, trim=0mm 50mm 0mm 0mm,clip]{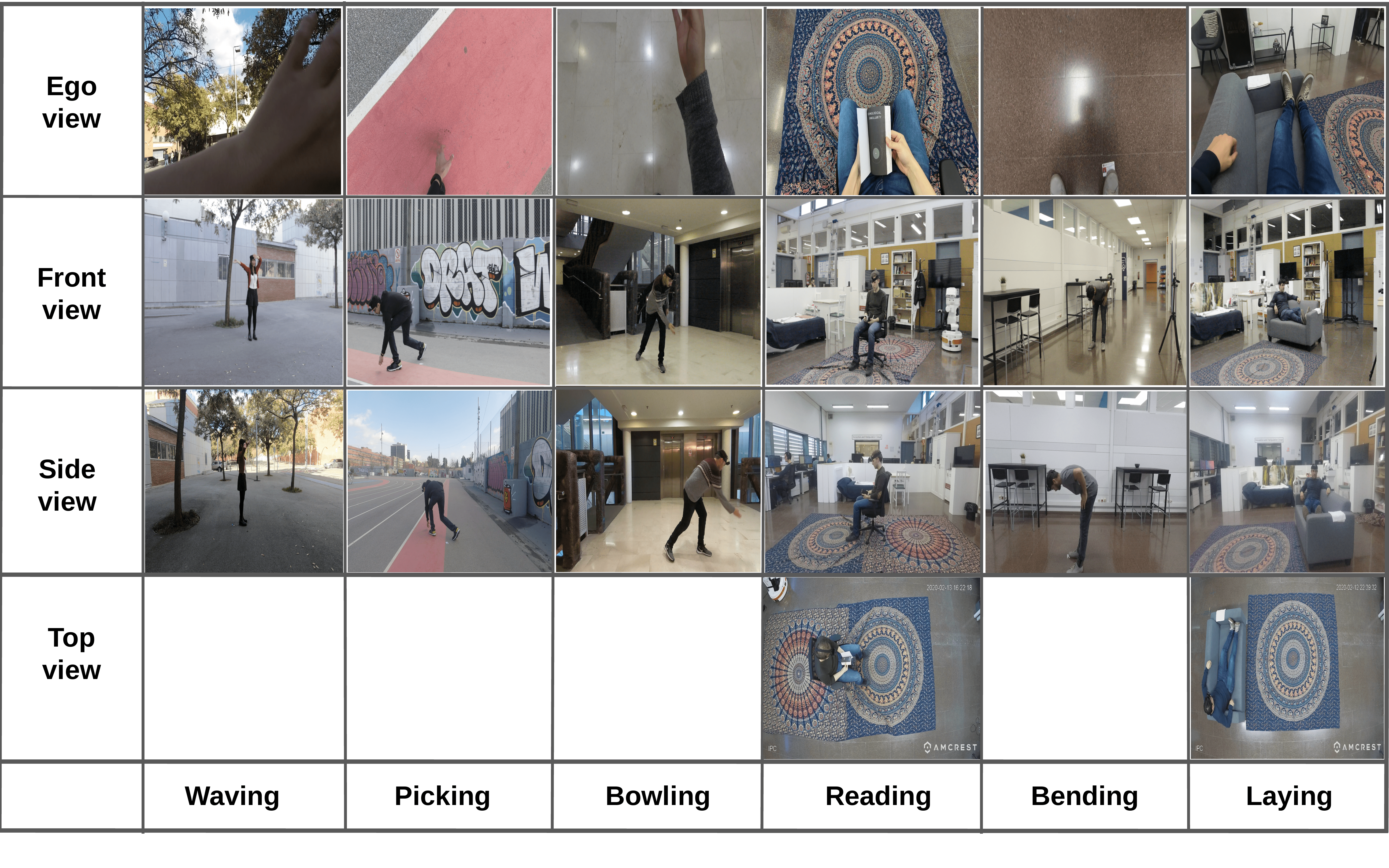}
\end{center}
\vspace{-4mm}
  \caption{Example of synchronized viewpoints in our dataset for different activities.}
  \label{fig:1st23rd_dataset}
\end{figure*}

\begin{table*}
\setlength{\tabcolsep}{4pt} 
\centering
\resizebox{\textwidth}{!}{%
\begin{tabular}{lcccccccccccccccc}
\toprule
Dataset & \#activities & \#videos & \#people & \#view & \#body joints & \#scenes & \#fps & \#camera loc. & \#duration & \#GT \\
\cmidrule(lr){1-1}  \cmidrule(lr){2-2} \cmidrule(lr){3-3} \cmidrule(lr){4-4}  \cmidrule(lr){5-5} \cmidrule(lr){6-6} \cmidrule(lr){7-7} \cmidrule(lr){8-8} \cmidrule(lr){9-9} \cmidrule(lr){10-10} \cmidrule(lr){11-11} 
MotionGraph \cite{jiang2017seeing} & 19 & 18 & 10 & 1 & 25 & 4 indoor & 30 & chest & 1-3m & KineticsV2 \\
You2me \cite{ng2020you2me} & 4 & 42 & 10 & 1 & 25 & indoor & 30 & chest & 2m & Kinect\&Panoptic \\
\cite{yuan2019ego}  & 8 & 24 & 5 & 2 & 22 & 2(indoor/outdoor) &30 & head & 20s & 3rd-view(2D) \\ 
{\bf Our} & {\bf 40} & {\bf 1950} & {\bf 14} & {\bf 4} & {\bf 17} & {\bf 4(indoor/outdoor)} & {\bf 25} & {\bf head} & {\bf 8-25s} &{\bf 3rd-view(3D)} \\
\hline
\end{tabular}
}
\vspace*{-.75em} 
\caption{Comparison of publicly available datasets for 3D egocentric pose estimation.}
\label{tab:comparisonDatasets1}
\vspace{0mm} 
\end{table*}

\begin{table*}
\setlength{\tabcolsep}{4pt} 
\centering
\resizebox{\textwidth}{!}{%
\begin{tabular}{lcccccccccccccccc}
\toprule
Dataset & \#activities & \#videos & \#people & \#view & \#scenes & \#fps & \#add.camera& \#duration & full pose visible& Task \\
 & & & & & & & location &  & on add. camera & 
\\
\cmidrule(lr){1-1}  \cmidrule(lr){2-2} \cmidrule(lr){3-3} \cmidrule(lr){4-4}  \cmidrule(lr){5-5} \cmidrule(lr){6-6} \cmidrule(lr){7-7} \cmidrule(lr){8-8} \cmidrule(lr){9-9} \cmidrule(lr){10-10} \cmidrule(lr){11-11} 
    \cite{chan2016recognition} & 12 & 20 & 11 & 3  & 3 & 60 & hands &  30s & - & gesture rec.\\
    \cite{soran2014action} &28  & 140 & 5 & 4 & 1(Lab) &30fps  & side,back,top & 1-2s & - & action rec.\\
   \cite{fan2017identifying}  & uncons. & 7 & 4 & 3 & 2 &30 & side & 5-10m & - & pers. disam. \\ 
   \cite{fan2017identifying}  & uncons. & 7 & 4 & 3 & 2 &30 & side & 5-10m & - & pers. disam. \\ 
 \cite{yonetani2016recognizing}  & 7 & 1226  &  6 & 2 & 6 &60 & head & 1.5s & -& act./react. \\
 \cite{bambach2015viewpoint}  & 4 & 24  & 4  &2  &  3& 30&  head& 90s & - & act.rec.\\
{\bf Our} & {\bf 40} & {\bf 1950} & {\bf 14} & {\bf 4} & {\bf 4(in/out)} & {\bf 25} & {\bf side,front,top} & {\bf 8-25s} & {\bf yes} &{\bf 3d pose est.} \\
\bottomrule
\end{tabular}
}
\vspace*{-.75em} 
\caption{Comparison of datasets combining first-person videos with other's viewpoint synchronized videos.}
\label{tab:comparisonDatasets2}
\vspace{0mm} 
\end{table*}

\section{First2Third-Pose Dataset}
\label{sec:dataset}

We next introduce \textit{First2Third-Pose}, a large dataset of short videos covering a variety  of  human  pose  types and including multi-third-person-views in addition to first-person view.  

\vspace{1mm}
\noindent{\bf Dataset collection.}
We built a multi-view synchronized dataset wherein we capture 14 people (in turns) of varying height, weight and genders while performing 40 different activities in both indoors and outdoors environments (lab, streets, parks, corridors, basketball courts, and parking). Every individual is asked to wear the head mounted camera which captures his/her egocentric view. We use the Go Pro Hero 4 in normal view setting which records in $1920 \times 1080$ resolution at 25 fps. All indoor and outdoor environments are equipped with two static cameras that capture the side and front view. We use the Go Pro Hero 3 in $1920 \times 1080$ resolution at 25 fps to record the side view and the Sony DSLR in $1920 \times 1080$ resolution at 25 fps to record the front view.  An example of outdoor capture setup is shown Figure \ref{fig:setup}. The `Lab' scene is equipped with an additional Amcrest camera that we use  under the wide setting with a frame rate of 20 fps and $2304\times 1296$ resolution. The top view captures a perspective from above the individual; the side view captures the 3rd person perspective from either left/right side recorded parallel to the individual; the front view captures the 3rd person view of the individual from the front; and the egocentric view captures the 1st person perspective of the individual. Each person performs about 40 activities in two indoor and two outdoor locations. 
We record a total of 1950 activity sequences lasting between 8 to 25 seconds, with a total duration of 2.5 hours. The activities include sports actions as boxing, basketball, soccer and day-to-day tasks like reading, typing or sitting on couch/chair/ground. Examples of synchronized viewpoints for different activities in our dataset are shown in Figure \ref{fig:1st23rd_dataset}.

\vspace{1mm}
\noindent{\bf Post-processing.} 
To enable the synchronization across multiple views, we recorded  one video for each view in a location (ego, top, front, side). Before starting to enact each activity, the participants are asked to clap. This clap is visually captured in egocentric camera and heard across all the multiple views. The sound of the clap is used to determine the staring point of each activity. We use the front views to check how long the activity has been performed. This time duration is noted and the video is manually scanned to find starting points of the activity. The activity is then clipped out from the main video and annotated using the name, activity class performed, location and the view that videos presents. Each video clip has one activity class. This is done  for all the front, ego, top and side views. 

\vspace{1mm}
\noindent{\bf Dataset annotation.}
The body pose is represented by $J=  17$ 3D joint positions.
Similarly to previous work on egocentric pose estimation \cite{yuan2019ego}, our ground truth is estimated from front views. However, instead of using only 2D joints as in \cite{yuan2019ego}, we first estimated 2D body poses by using Detectron2 \cite{wu2019detectron2}, and then we obtained 3D body estimations via a pre-trained lifting model \cite{pavllo:videopose3d:2019}. As the latter method reports an average error of 4-5 cm on large-scale datasets, we can assume that at most the same holds in our case. Visual inspection of 3D pose predictions on the test set corroborated the plausibility of such assumption.
It is worth to remark that as our dataset was captured in multiple scenarios, the camera extrinsic parameters are not available. Therefore, the computation of the ground truth could not benefit from triangulation methods by using multiple views. However, the ground truth annotations are not required to compute the joint embedded space we exploit in the proposed approach. 

\vspace{1mm}
\noindent{\bf Dataset comparisons.}
In Table \ref{tab:comparisonDatasets1} we summarize the characteristics of our proposed dataset with respect to other available benchmarks for 3D egopose estimation. It can be observed that our dataset scales existing ones in terms of number of videos and presents more variability in terms of background scenes, number of participants and activities.  Furthermore, it provides multiple views of the same scene (egocentric, top, side, front). Therefore, it is currently  the largest and comprehensive dataset for 3D egopose estimation from videos.

In Table \ref{tab:comparisonDatasets2} we compare qualitatively existing datasets with syncronized videos including at least one first-person viewpoint, and we show that our dataset is the only one suited for the task of 3D egopose estimation. 
We stress that while other paired datasets with first- and third-person views currently exists, e.g. \cite{sigurdsson2018actor}, we did not include them in  Table \ref{tab:comparisonDatasets2} since they are not synchronized and therefore not suitable for our task. All our data will be made publicly available upon acceptance.

\begin{figure*}
\begin{center}
  \includegraphics[width=\linewidth]{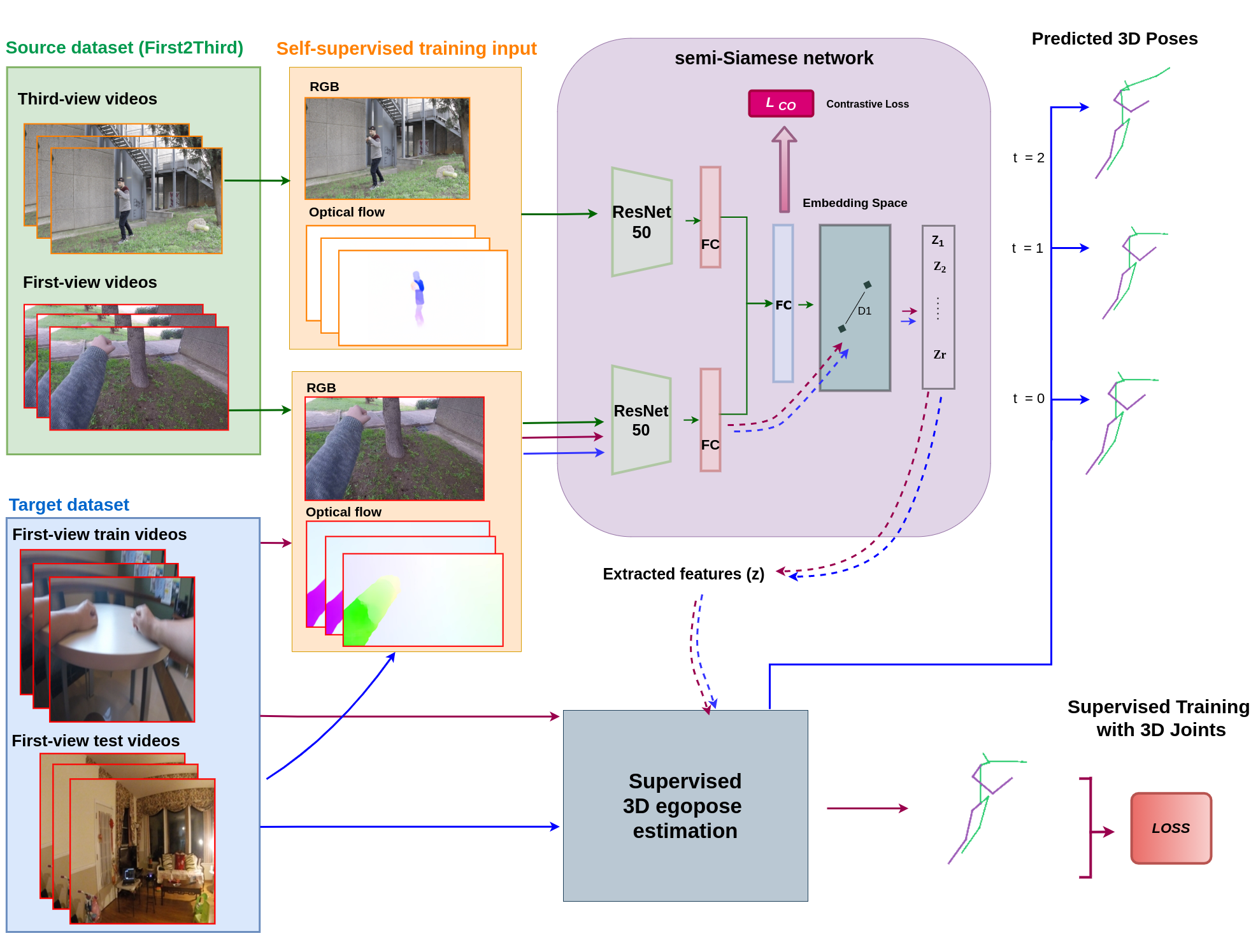}
\end{center}
\vspace{-4mm}
\caption{Our model uses a semi-Siamese architecture to learn to detect if a pair of first- and third-view videos of the \textit{First2Third-Pose} paired source dataset are syncronized or not, by minimizing a contrastive loss (\textcolor{ao(english)}{green} arrows). 
This pretext task leads to learn a joint embedding space, where the gap between the first-view and third-view worlds is minimized. The so learned joint embedding space can in principle be leveraged by any supervised method for 3D egopose estimation on a target dataset, without a need for domain adaptation. At both train time (\textcolor{brown(web)}{brown} arrows)) and test time (\textcolor{blue}{blue} arrows), the semi-Siamese network is used for feature projection onto the learned joint embedded space. $z$ is 64-
dimensional vector, obtained once removed the softmax layer of the Siamese network pre-trained with our dataset.}
\label{fig:model}
\end{figure*}

\section{Approach}
\label{sec:approach}
Given an egocentric video sequence as input, our goal is to estimate the 3D body joints of the camera wearer as output. More formally, for each egocentric video frame at time $t$, the output is a set of $J$ joint 3D coordinates corresponding to the skeleton of camera wearer at frame $t$, say $p_t \in {R}^{3J}$.  

Our key insight is to build image features allowing to discriminate different 3D human poses by projecting and aligning data from the first and third-view onto a shared representation space.
Formally, our objective is to learn functions $f_1: \mathcal{R}^F \rightarrow \mathcal{R}^J$ and $f_2: \mathcal{R}^T \rightarrow \mathcal{R}^J$ which map first and third views corresponding to the same 3D pose respectively onto nearby points in a joint embedded space.
Positive first-third pairs are extracted from synchronized videos and fed into a two-stream Siamese architecture with a first view subnetwork and a third-view subnetwork, each producing 64-D embeddings. A curriculum-based mining schedule is used to select appropriate negative pairs which are then trained using a contrastive loss, as detailed below. Our contrastive loss evaluates if pairs of first- and third- person views are syncronized. To solve such syncronization task, the network must learns what these extremely different views have in common when they are syncronized: the 3D human body pose of the person visible in the third-view but invisible in the first-view.
Consequently, the proposed syncronization pretext task translates well to the downstream task of 3D egopose estimation.

 \subsection{Learning a joint first/third view embedding}

First  and third-person perspectives are very different in appearance (see Figure \ref{fig:1st23rd_dataset}). However, previous work \cite{fan2017identifying,sigurdsson2018actor} has shown that it is possible to find spatial and motion domain feature correspondences between video types in a self-supervised fashion. While this has proved to be useful in the context of activity recognition and second-person disambiguation, its usefulness for 3D body pose estimation has never been investigated before. 
Our goal is therefore to attain a semantically meaningful space where paired first- and third-person video are close to each other while proximity of unpaired videos is avoided. 
The pretext task is to classify paired views into synchronized and unsynchronized, which in turn corresponds to determine if pairs of  first-person and a third-person views correspond to the same 3D human pose.
Similarly to \cite{fan2017identifying}, this is achieved by training  a two-streams semi-Siamese Convolutional Neural Network (CNN) with a ResNet50 backbone \cite{he2016deep}. Differently from \cite{fan2017identifying}, where self-supervision is performed separately on the the spatial and motion domains, we treat them jointly and we minimize a single contrastive loss. 
Since first- and third- person views are very different in appearance, parameter sharing is allowed only for the last fully connected layer. The semi-Siamese architecture is shown on the upper part of Figure \ref{fig:model}. 

At training time, we feed to the network paired exemplars of first-view and third-view, consisting of stacked RGB and optical flow frames. To estimate optical flow on your dataset, we used the FlowNet2 \cite{Ilg17} architecture pre-trained on the FlyingChairs and subsequently fine-tuned on the Things3D datasets. Specifically, for each given video frame, we computed the optical flow field as a forward pass for a window of length 11 centered on it. We corroborated the quality of the estimation by visual inspection of the results.
We then minimize a contrastive loss measuring the Euclidean distance for
positive exemplars and a hinge loss for negative ones. Our loss function is as follows:
$$
 L = \sum_j^M y_j||x_j^f - x_j^t||^2 + (1-y_j)\max(0, m -||x_j^f - x_j^t||)^2
$$
where $M$ is the number of frames in a batch, $m$ is a predefined constant margin, $x^f$ and $x^t$ indicate first and third views (stacked RGB and optical flow frames) respectively, $y_i$ is an indicator that takes value 1 if $x^f_j$ and $x^t_j$ are synchronized exemplars, and 0 otherwise.
Following its effectiveness in self-supervised learning as alternative to random exploration, we adopted a curriculum learning strategy for training \cite{bengio2009curriculum}. Concretely,  we defined as \textit{easy negatives} pairs of first- and third video clips corresponding to a same person doing a different activity in the same environment. 
\textit{hard negatives} were defined as pairs of first- and third video clips corresponding to a same person doing the same action in the same environment at different intervals of time.

In the following subsection, we aim at leveraging this joint representation space learned with our source \textit{First2Third-Pose} dataset to estimate the 3D body pose of the person behind the camera in first-view videos from a target dataset.

\subsection{Transfer to 3D Egopose estimation}

As  a byproduct of learning to classify paired first- and third-view videos into
synchronized and unsynchronized, the representation gap between the two perspective
views is minimized. Therefore, the representation space shared by the first and third-view enables to transfer a first-view projection onto this space, which should be similar to a paired third-view projection, to 3D human pose regression via supervised learning.

We used the semi-Siamese network pre-trained on the pretext task with the source \textit{First2Third-Pose} dataset to extract features, which are then useful for 3D pose estimation from egocentric videos. More specifically, stacked RGB and optical flow frames from the unpaired egocentric video are fed into the first-view stream of the network in a forward pass. 
The bottom part of Figure \ref{fig:model} shows how target train and test egocentric video datasets can be used as input to our pre-trained semi-Siamese network to extract features that are subsequently used as additional features by a supervised 3D egopose estimation method. 
The network can in principle be used as feature extractor by an arbitrary supervised model for 3D egopose estimation, at both training and test time.
At training time, the target dataset is used as input for both the supervised method at hand and the pre-trained Siamese network. However, the latter is used only to perform a forward pass allowing to project the input videos into the shared representation space and hence obtain discriminative features used as additional input for the supervised approach.
Through experiments, we will show that even if the embedding space has been learned relying on the \textit{First2Third-Pose} source dataset, it transfers well on a different target egocentric video dataset, without a need for domain adaptation.

\section{Experiments}
\label{sec:experiments}

\vspace{1mm}
\noindent{\bf Implementation details.} We train a semi-Siamese network based on the ResNet50 backbone that takes as input on each stream an RGB frame stacked to the optical flow fields for a set of 10 consecutive frames. We used FlowNet2 \cite{Ilg17} to estimate optical flow. The output of the ResNet in each stream is fed to a fully connected layer of dimension $100$. The last common fully connected layer has dimension $64$.

We generate the training data by splitting our Multiview dataset into 150k training frames and 40k testing frames. The train set includes activities performed by 10 people (8 actors, 2 actresses), while the test set includes activities performed by 4 unseen people (2 actors, 2 actresses).
To train the Siamese Network we generate positive and negative image pairs. The positive pairs are generated taking synchronized first- and third- view (front) video frames. As we adopted a curriculum learning strategy for training, we generated negative pairs in two ways. Easy negative pairs correspond to first- and third-view videos of same person doing different activities in the egocentric and front images, but in the same environment. Hard negative pairs correspond to shifted time intervals in paired first-and ego views. We follow curriculum learning to train the Siamese network for 2 epochs. We train the network using easy negative pairs for first epoch and use the hard negative pairs for the second epoch. We use the contrastive loss with a margin of 0.9 to train the network using these pairs. Training time is 96 hours on a single GPU for 2 epochs. The learning rate is set to 0.0001 with the momentum 0.9 and the weight decay  5e-4. Each predicted 3D body pose has the hip joint positioned at the origin of the coordinates system.
The first axis is parallel to the ground and points to the wearer’s facing direction. The second axis is parallel to the ground and points to the left hip. The third axis is perpendicular to the ground and points in the direction of the spine. To account for people size variability, we normalize each skeleton for scale based on the individual's shoulder width.

\begin{table*}[]
\resizebox{\textwidth}{!}{%
\begin{tabular}{l c c c c c c c c c c c c  }
\toprule
 & \multicolumn{12}{c}{First2Third Dataset}\\
\cmidrule(lr){2-11} \cmidrule(lr){11-12} \cmidrule(lr){13-13}
& Hip  & Neck & Head & Shoulders & Elbows & Wrists & Thorax & Knees  & Feet & UppBody & LowBody & Avg \\
\cmidrule(lr){1-1} \cmidrule(lr){2-11} \cmidrule(lr){11-12} \cmidrule(lr){13-13}

MotionGraph \cite{jiang2017seeing} & 3.40 & 8.03 & 12.40 & 7.74 & 20.93 & 39.13 & 10.73 &32.50 & 53.81  & 16.98  & 25.63 & 20.54 \\
{\bf MotionGraph-SS} &  \textbf{3.37} & \textbf{5.96 } & \textbf{9.57} & \textbf{6.23} & \textbf{18.86 }& \textbf{31.09} & \textbf{8.38} & \textbf{28.66 } & \textbf{45.14} & {\bf 13.89} & {\bf 22.04}  & {\bf 17.25 }  \\ 
\cmidrule(lr){1-1} \cmidrule(lr){2-11} \cmidrule(lr){11-12} \cmidrule(lr){13-13}

 you2me  \cite{ng2020you2me}&  2.96 & 10.45 & 15.31& 9.81 & 20.15 & 34.52 & 13.13 & 26.31  & 48.25 & 17.78  & 23.16 & 20.00 \\
{\bf you2me-SS}  & 2.99 & \textbf{8.66} & \textbf{13.26}& \textbf{8.23 }& \textbf{18.90 }& \textbf{33.61} & \textbf{11.34} & 27.08 & \textbf{47.74 }& {\bf 15.63}  & {\bf 21.33}  & {\bf 18.03}\\ 
\cmidrule(lr){1-1} \cmidrule(lr){2-11} \cmidrule(lr){11-12} \cmidrule(lr){13-13}

Deconvnet \cite{xiao2018simple}&  3.15 & 5.72 & 8.96& 5.77 & 17.13 & 31.26 & 7.81 & 27.84 & 45.48 & 13.35 & 21.85  & 16.85 \\
{\bf Deconvnet-SS} &  \textbf{3.12} & \textbf{5.69} & 9.30 & \textbf{5.71} & \textbf{14.89} & \textbf{27.10} & 8.03 &\textbf{24.00}  & \textbf{40.09} & \textbf{12.09} & \textbf{18.64}  & \textbf{14.78}  \\ 
\bottomrule
\end{tabular}
}
\caption{Average joint error in the \textit{First2Third} Dataset, in cm. 
}
\label{tab:results1}
\end{table*}

\begin{table*}[]
\resizebox{\textwidth}{!}{%
\begin{tabular}{l c c c c c c c c c c c c c }
\toprule
& \multicolumn{12}{c}{Invisible Poses Dataset~\cite{jiang2017seeing}} \\
\cmidrule(lr){2-11} \cmidrule(lr){12-13} \cmidrule(lr){14-14}
& Hip  & Neck & Head & Shoulders & Elbows & Wrists & Hands & Knees & Ankle & Feet & UppBody & LowBody & Avg \\
\cmidrule(lr){1-1} \cmidrule(lr){2-11} \cmidrule(lr){12-13} \cmidrule(lr){14-14}
MotionGraph \cite{jiang2017seeing} & 2.24  & 19.40 & 21.60 & 16.23 & 17.06 & 24.02 & 27.27 & 24.78 & 32.29 & 34.13 & 22.09 & 20.76 & 21.61\\
{\bf MotionGraph-SS} \cite{jiang2017seeing} & \textbf{2.14}  & \bf{17.20} & \bf{19.50} & \textbf{14.23}&	\textbf{14.81}&\textbf{	21.29}&\textbf{	24.32}&\textbf{23.32}&	\textbf{30.43}& \textbf{32.29} & {\bf 19.65}   &  {\bf 19.60}    & {\bf 19.64}  \\
\cmidrule(lr){1-1} \cmidrule(lr){2-11} \cmidrule(lr){12-13} \cmidrule(lr){14-14}
you2me \cite{ng2020you2me} & 2.14  & 15.50 & 17.10 & 15.17 & 19.34 & 28.54 & 32.23 & 22.48 & 33.94 	& 36.63 & 24.53 & 21.55 & 23.38  \\
{\bf you2me-SS}  & \textbf{1.97}  & \bf{15.00} & \bf{17.00} &  \textbf{13.87}	&\textbf{16.19}&	\textbf{24.04}	& \textbf{27.19}& 23.27	& \textbf{32.80} &	\textbf{35.37} & \textbf{20.94} &  \textbf{20.75}  & \textbf{20.87} \\
\cmidrule(lr){1-1} \cmidrule(lr){2-11} \cmidrule(lr){12-13} \cmidrule(lr){14-14}
Deconvnet~\cite{xiao2018simple}& \textbf{2.58} & 17.70 & 21.30 &  12.65 &	13.81&	20.87&	23.46&	22.00	&27.30&	29.04 & 18.46 & 17.98 & 18.29  \\
{\bf Deconvnet-SS}  & 2.64  & \bf{16.00} & \bf{18.80} &  \textbf{12.48}&	\textbf{13.59}&	\textbf{20.31}&	\textbf{22.82}&	\textbf{21.49}&\textbf{	26.20}&	\textbf{27.74}& \textbf{18.07} & \textbf{17.31} & \textbf{17.85}  \\
\bottomrule
\end{tabular}
}
\caption{Average joint error in the Invisible Poses Dataset~\cite{jiang2017seeing}, in cm.}
\label{tab:results2}
\end{table*}

\vspace{1mm}

\begin{figure*}[!t]
\begin{center}
  \includegraphics[width=\linewidth, height = 9cm, clip = true]{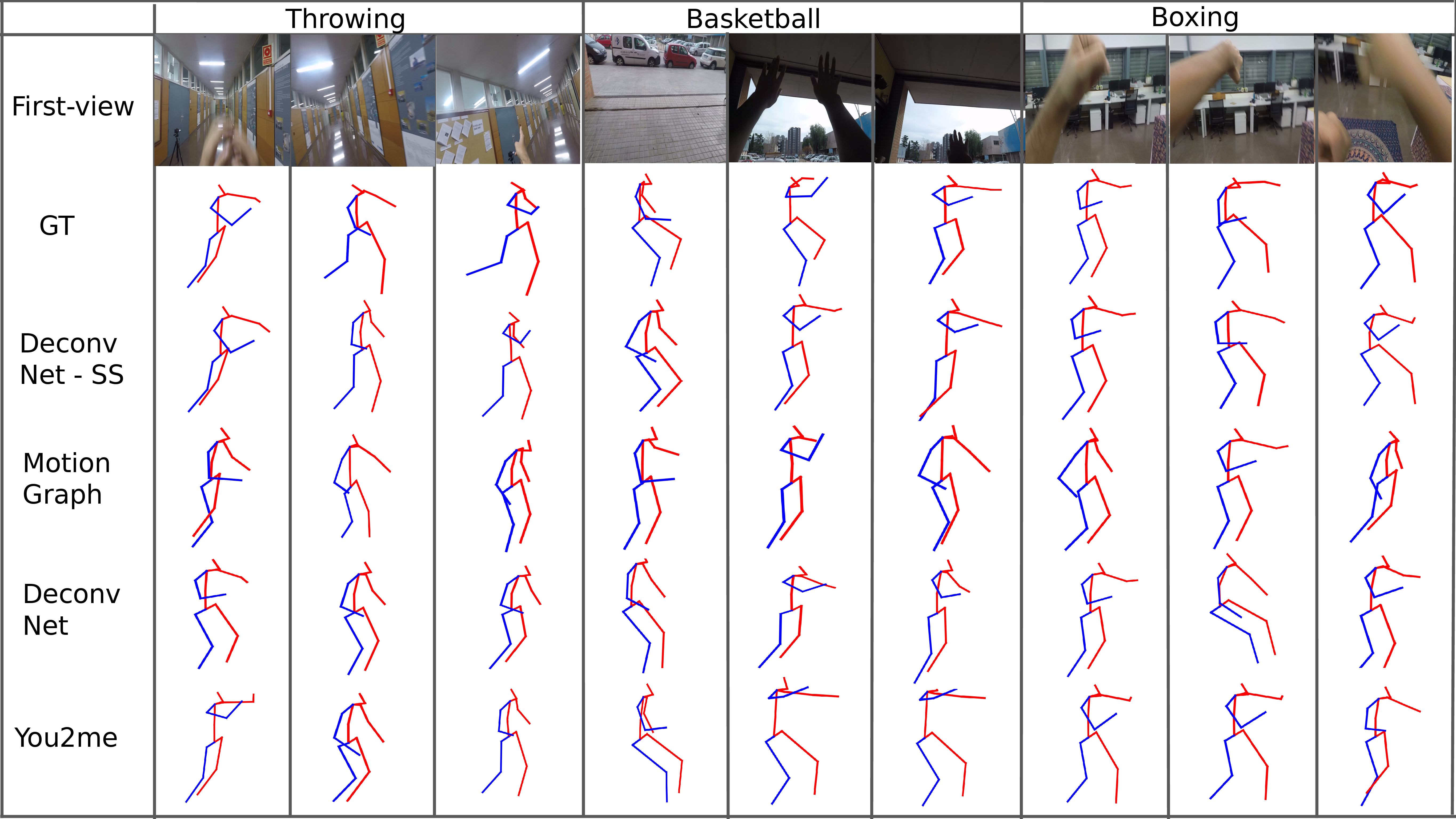}
\end{center}
\caption{Visual comparisons of predicted skeletons for three different activities. GT: ground truth. DeconvNet-SS:
proposed method. MotionGraph: state-of-the-art \cite{jiang2017seeing}. DeconvNet: end-to-end baseline \cite{xiao2018simple}. you2me: \cite{ng2020you2me} baseline. Test videos are from the \textit{First2Third-Pose} dataset test split. } 
\label{fig:vis_res}
\end{figure*}

\begin{figure*}[!t]
\begin{center}
  \includegraphics[width=\linewidth]{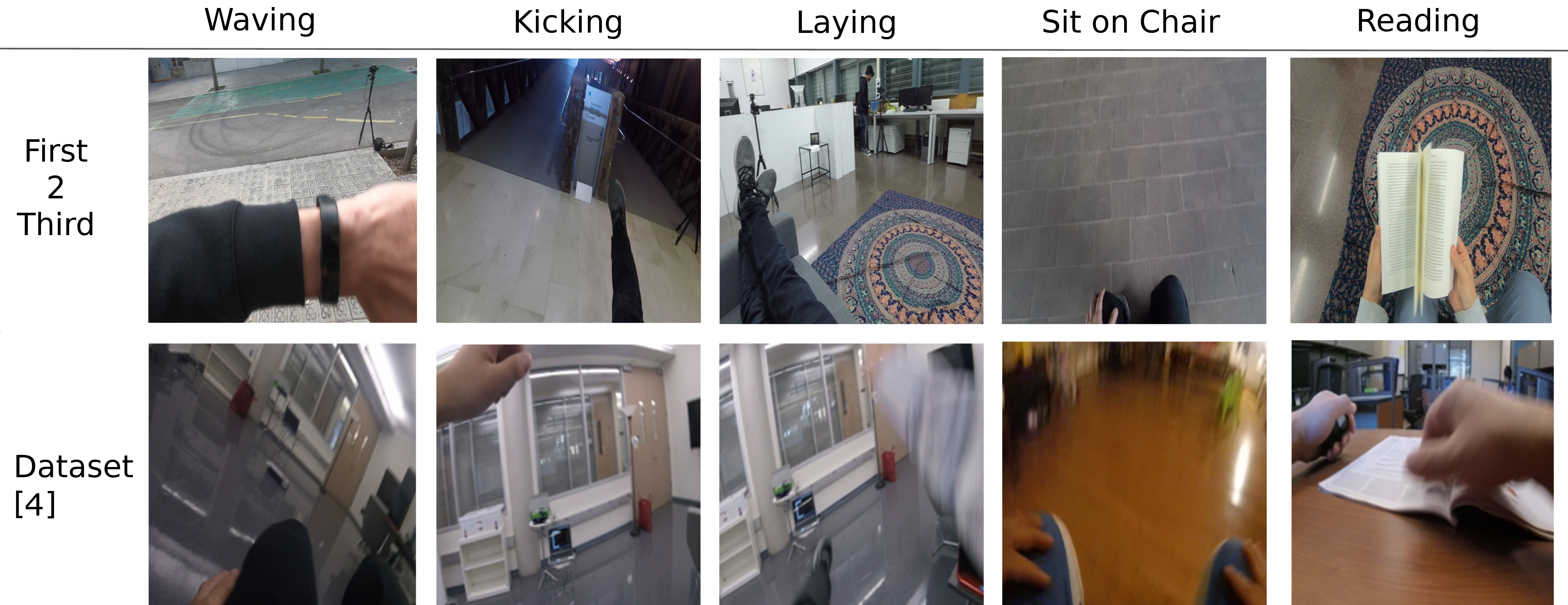}
\end{center}

\caption{Examples of activities captured in the two datasets: \textit{First2Third-Pose} (top) and Invisible Pose Dataset \cite{jiang2017seeing} (bottom).}
\label{fig:dataset_diff}
\end{figure*}

\noindent{\bf Datasets.} In addition to our \textit{First2Third-Pose} dataset described in Section \ref{sec:dataset}, we use the dataset introduced in  \cite{jiang2017seeing}. The first-view for the two datasets has been captured wearing the camera on the head for our dataset and on the chest for the other.  More details about the dataset   \cite{jiang2017seeing} can be found in Table \ref{tab:comparisonDatasets1}. Difference in appearance with our source \textit{First2Third-Pose} dataset can be appreciated in Figure \ref{fig:model} (see source and target first-view videos).
Figure \ref{fig:dataset_diff} illustrates the difference in appearance between two datasets used in the paper for a same activity: our proposed \textit{First2Third-Pose} captured by a head mounted camera, and the Invisible pose dataset \cite{jiang2017seeing} captured by a chest mounted camera. 

\begin{figure*}[!t]
\begin{center}
  \includegraphics[width=0.9\linewidth, height = 9cm, clip = true]{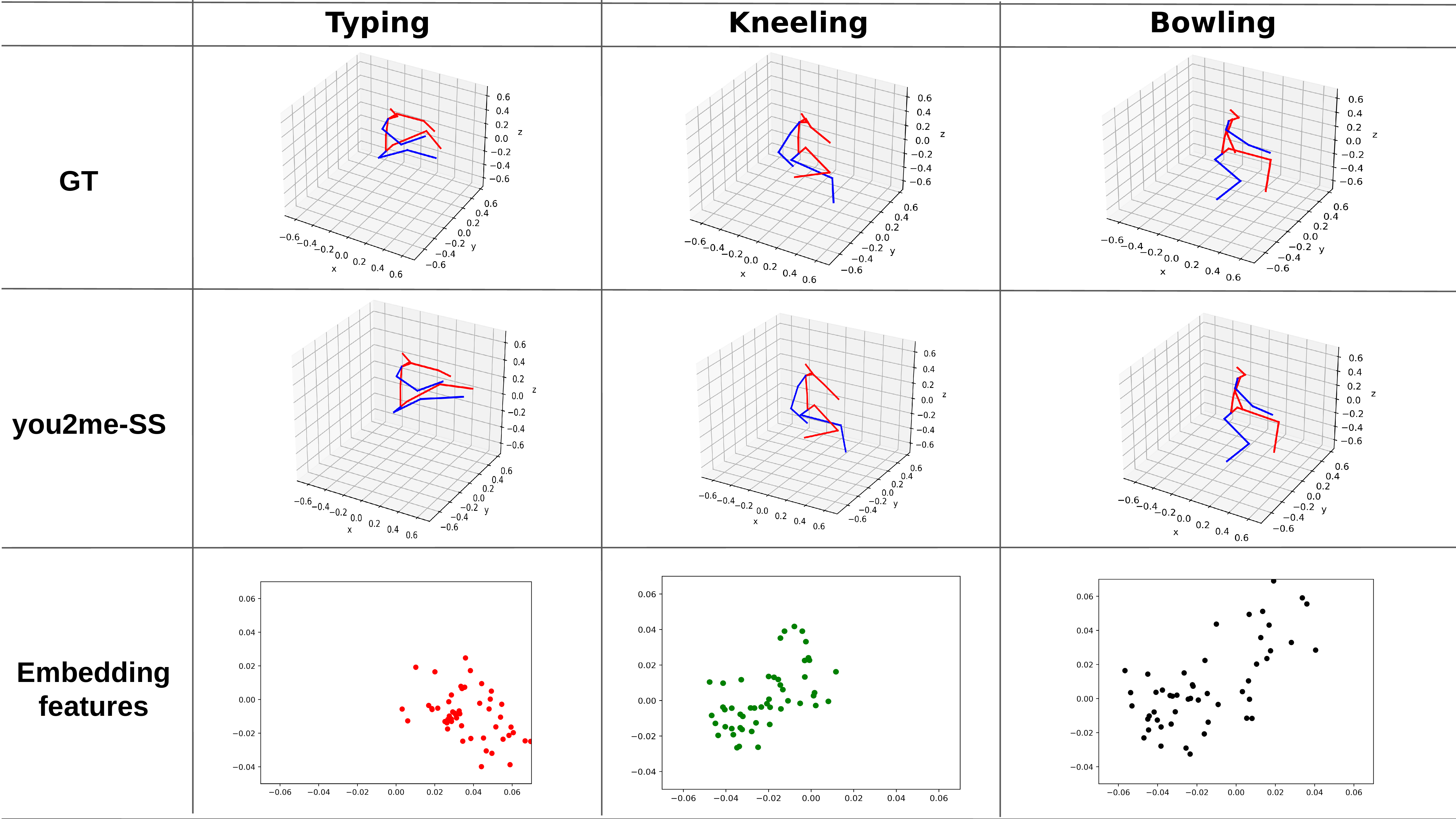}
\end{center}
\caption{Scatter plots of embedding features for three different 3D body poses. Similar body poses have similar features.} 
\label{fig:vis_feat}
\end{figure*}


\vspace{1mm}
\noindent{\bf Baselines.} 
We considered two state-of-the-art methods for 3D egocentric pose estimation \cite{jiang2017seeing,ng2020you2me}, and we considered also a baseline method tailored for 3D pose estimation from third-view videos \cite{xiao2018simple} that we adapted to our task. {\bf MotionGraph} \cite{jiang2017seeing} is currently the state-of-the art method for predicting 3D body pose from real egocentric videos without a second interacting person. We used the publicly available author's code \footnote{http://www.hao-jiang.net/egopose/index.html} to extract static and motion features from our dataset, and modified the MotionGraph dynamic programming algorithm to account for the features extracted by relying on our joint embedded space. We retrained the model for both datasets, using the 300 quantized poses as used in \cite{jiang2017seeing}.
{\bf You2me} \cite{ng2020you2me} has been recently proposed as a method able to account for the visible second person interacting with the camera wearer, as it leverages his/her 3D pose estimates to improve egopose predictions. Even if there are not second persons in our \textit{First2Third-Pose} dataset, this approach is still a valid alternative to state-of-the-art MotionGraph, since the use of a recurrent long short-term memory (LSTM) network ensures smooth frame to frame 3D body pose transitions. 
We used the authors's code \footnote{https://github.com/facebookresearch/you2me}, that takes as input motion and appearance based features, and used additional features vector extracted leveraging our learned joint embedding as input to the LSTM for each frame. We trained this model for both datasets, using 700 quantized poses for upper body and 100 for lower body, as in \cite{ng2020you2me}.
We also adapted and trained from scratch the 3D human pose estimation baseline method proposed in \cite{xiao2018simple}\footnote{https://github.com/una-dinosauria/3d-pose-baseline}, {\bf DeconvNet}, that adds  deconvolutional layers to ResNet. We altered the output space for 3D joints and minimized the mean-squared error on the training set. We found this off-the-shelf deep pose method extremely effective for ego-pose estimation, specially on our \textit{First2Third-Pose} dataset that, being large scale, is well suited for end-to-end learning.

\vspace{1mm}
\noindent{\bf Evaluation metric.}
Each skeleton is rotated so the shoulder is parallel to the yz plane and the body center is at the origin.
The  error  is then computed as  the  Euclidean  distance  between the predicted 3D joints and the ground truth, averaged over the full sequence and scaled to centimeters based on a reference shoulder distance of 30 cm.

\vspace{1mm}
\noindent{\bf Results.}
In Table \ref{tab:results1} and Table \ref{tab:results2} we show the average error per joint and for all joints (in cm) obtained on the test set of our dataset and on the dataset \cite{jiang2017seeing} respectively. We denote by MotionGraph-SS, you2me-SS and Deconvnet-SS our Self-supervised approach based on the methods MotionGraph \cite{jiang2017seeing}, you2me \cite{ng2020you2me} and DeconvNet \cite{xiao2018simple}, respectively.
In addition, following competitive methods \cite{ng2020you2me,jiang2017seeing}, we present results in terms of errors averaged separately for the upper body (Neck, Head, Thorax, Spine, Shoulders, Elbows, Wrists, Hands) and
lower body joints (Hips, Knees, Ankles, Feet).  
Overall, these tables show that leveraging features extracted from the common representation space learnt with our proposed dataset consistently gives better results over existing supervised methods for 3D egopose estimation. The fact that our approach results effective also on the dataset \cite{jiang2017seeing}, demonstrates that the features extracted relying on the learned joint embedding space can be efficiently transferred to arbitrary egocentric videos, without the need of domain adaptation.

Figure \ref{fig:vis_res} shows qualitatively that using features extracted by leveraging the learned joint embedding space indeed gives better results.

In Table \ref{tab:timeComplexity}, we also considered the time complexity of the proposed approach with respect to the methods we compare to, measured in terms of Floating Point Operations per Second (FLOPS) computed by using a dedicated PyTorch library  \footnote{ https://pypi.org/project/ptflops/} for the neural network based models. For Motion graph, whose dynamic programming code released by the authors only includes vector-to-vector operations,  we report the FLOPS corresponding to the sum of these operations during the execution of the program. As it is standard when computing FLOPS, we assume that all input features have been pre-computed and loaded. As the size of the 
 self-supervised features $\textbf{z}$ is relatively small compared to the rest of input features, its effect on the FLOPS counting is negligible for the considered methods. 
However, given that precisely the same number of FLOPS may have radically different run-times, we also report the run-times for training and inference. We performed all the experiments on a single  workstation equipped with an Intel(R) Xeon(R) CPU E5-2620 v3 @ 2.40GHz, 128GB RAM 2133 MHz, NVIDIA Tesla K40c with 2880 CUDA core and operating system Ubuntu 14.04. 
It can be observed that, for the methods based on neural networks, while the run time at training time is significantly increased, at inference time only a very little increment is observed for getting the projection of the ego-view on the shared representation space. Therefore, at inference time, the proposed approach can be considered suitable for real-time applications.
\begin{table*}
\setlength{\tabcolsep}{4pt} 
\centering
\resizebox{\textwidth}{!}{%
\begin{tabular}{lcccccc}
\toprule
Method & Tr. time (h/epoch) & Inf. time (sec/image) & FLOPS (GMac) \\
\cmidrule(lr){1-1}  \cmidrule(lr){2-2} \cmidrule(lr){3-3} \cmidrule(lr){4-4}   
MotionGraph \cite{jiang2017seeing} & 0.10 & 0.01 & 0.45 \\
You2me \cite{ng2020you2me} & 2.50 & 0.17 & 12.74  \\
DeconvNet \cite{xiao2018simple}  & 1.24 & 0.28 & 9.95  \\ 

MotionGraph-SS  & 0.25 & 0.01 &  0.65 \\
You2me-SS & 26.06 & 0.30 &  12.74  \\
DeconvNet-SS   & 13.00 & 0.49 &  9.95 \\

Ours &  40.33 &  0.06 & 9.82  (Tr.)/4.91 (Inf.) \\

\hline
\end{tabular}
}
\vspace*{-.75em} 
\caption{Comparison of time complexity (FLOPS) and actual run-times for training one epoch on the \textit{First2Third-Pose} dataset and for inference on a single image.}
\label{tab:timeComplexity}
\vspace{0mm} 
\end{table*}

\vspace{1mm}
\noindent{\bf Insights on the joint embedded space.}
To verify that the features obtained using our embedded space, say embedded features, are discriminative for 3D egopose estimation, we first apply PCA to the
feature matrix to reduce the feature dimension to two, and then we visualized the features of videos corresponding to different activities
via a scatter plot (each dot is a frame).
In Figure \ref{fig:vis_feat} we show example poses for three different activities (ground truth and prediction) together with the scatter plot of corresponding embedding features for the surrounding 1 second video segment. 
 The 3D skeleton corresponding to the activity \textit{typing} differs from both  \textit{bowling} and \textit{kneeling}, while those of \textit{bowling} and \textit{kneeling} are more similar. This is also reflected by the corresponding features.

\section{Model interpretation}
\label{sec:interpretability}
\begin{figure}[!t]
 \centering
    \subfloat[\centering   ]{\includegraphics[width=5.5cm]{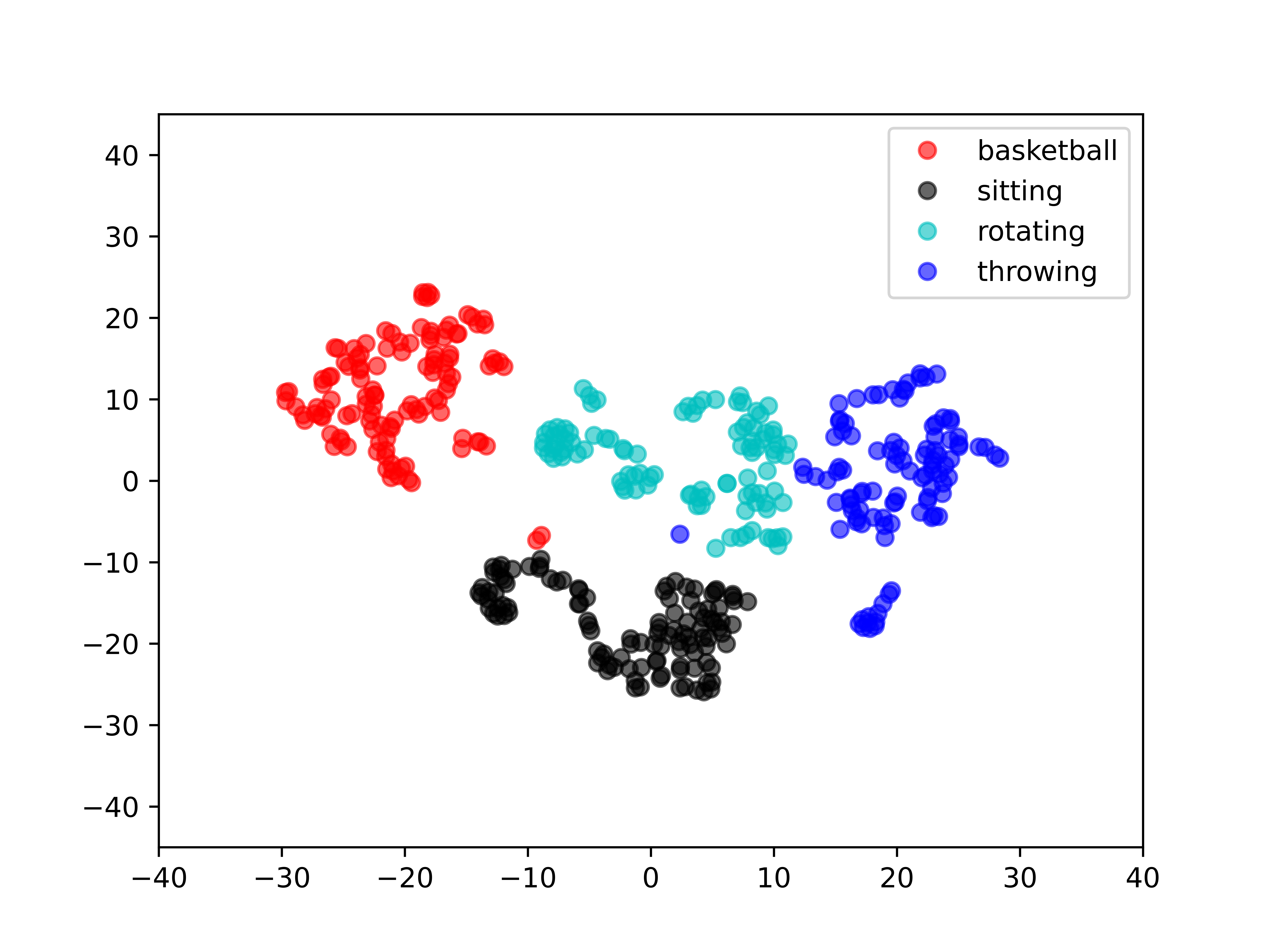}}%
    \subfloat[\centering  ]{\includegraphics[width=5.5cm]{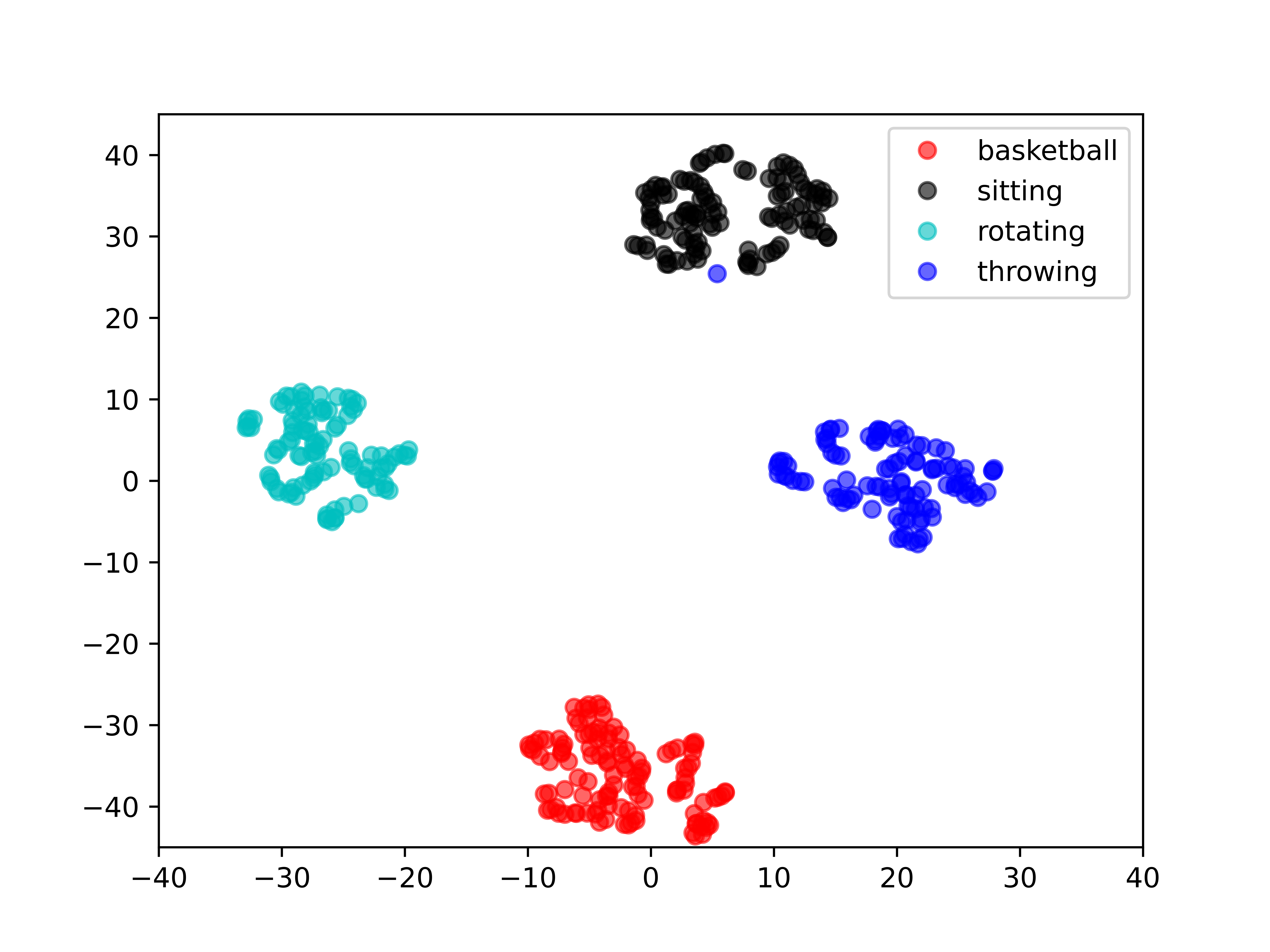} }%
\caption{t-SNE of four different activities captured from first- (a)  and third-person (b) views.} 
\label{fig:t-SNE}
\end{figure}

To shed light on the structure of the learned embedded space, we evaluated numerically the distance between corresponding first and third view embeddings of a same action. In addition, we visualized the embedding vectors corresponding to different activities captured from the same point of view (first-or third).
In Figure \ref{fig:t-SNE} we visualize via t-SNE the 1st- and 3rd-views embedding relative to 4 different actions in the joint space (each dot is a video). In both domains, the embedding of different activities are well separated, meaning they are discriminative for activity recognition. This also suggests that the embedded space can be useful for egocentric action recognition.

Furthermore, we examined the relationship between first-view and third-view feature projections in our embedding space by using Canonical Correlation Analysis (CCA) \cite{hotelling1992relations}. To make the results more easily interpretable, we grouped the set of activities captured by our dataset into four classes, depending on the type of body movement: 1) short actions involving the whole body (rotating, walking, jogging, etc.); 2) actions requiring a vertical movements of the body or body parts (crouching, sit on ground, pick object right,etc.);
3) short actions requiring mainly the movement of hands or feet (throwing, bowling, etc.);
4) activities made of a sequence of several short actions (basketball, exercise, etc.).  Table \ref{tab:CCA} reports the CCA coefficient among first-third view pairs of embeddings for each of these four classes. The correlation among first- and third- view embedding is strong, meaning that the features encode relevant geometrical information linking the two views.

\begin{table*}[]
\resizebox{\textwidth}{!}{%
\begin{tabular}{l c c c c}
\toprule
 & \multicolumn{4}{c}{Class based Activities}\\
 \cmidrule(lr){2-5} \cmidrule(lr){1-1}
\diagbox{Top}{Front}
&Complex&Hands \& Feet&Vertical Movement&Whole Body \\
 \cmidrule(lr){2-5} \cmidrule(lr){1-1}
Complex& {\bf 0.58} &0.15&-0.07&0.01 \\
Hands \& Feet&0.11&{\bf 0.62} &-0.01&0.00\\
Vertical Movement&-0.06&-0.08& {\bf0.43} &-0.01\\
Whole Body&0.04&-0.06&0.07&{\bf 0.52} \\
\bottomrule
\end{tabular}
}
\caption{Average CCA coefficient among first and third view pairs of embeddings for four different classes of activities.}
\label{tab:CCA}
\end{table*}

\begin{figure*}[!t]
\begin{center}
  \includegraphics[width=\linewidth]{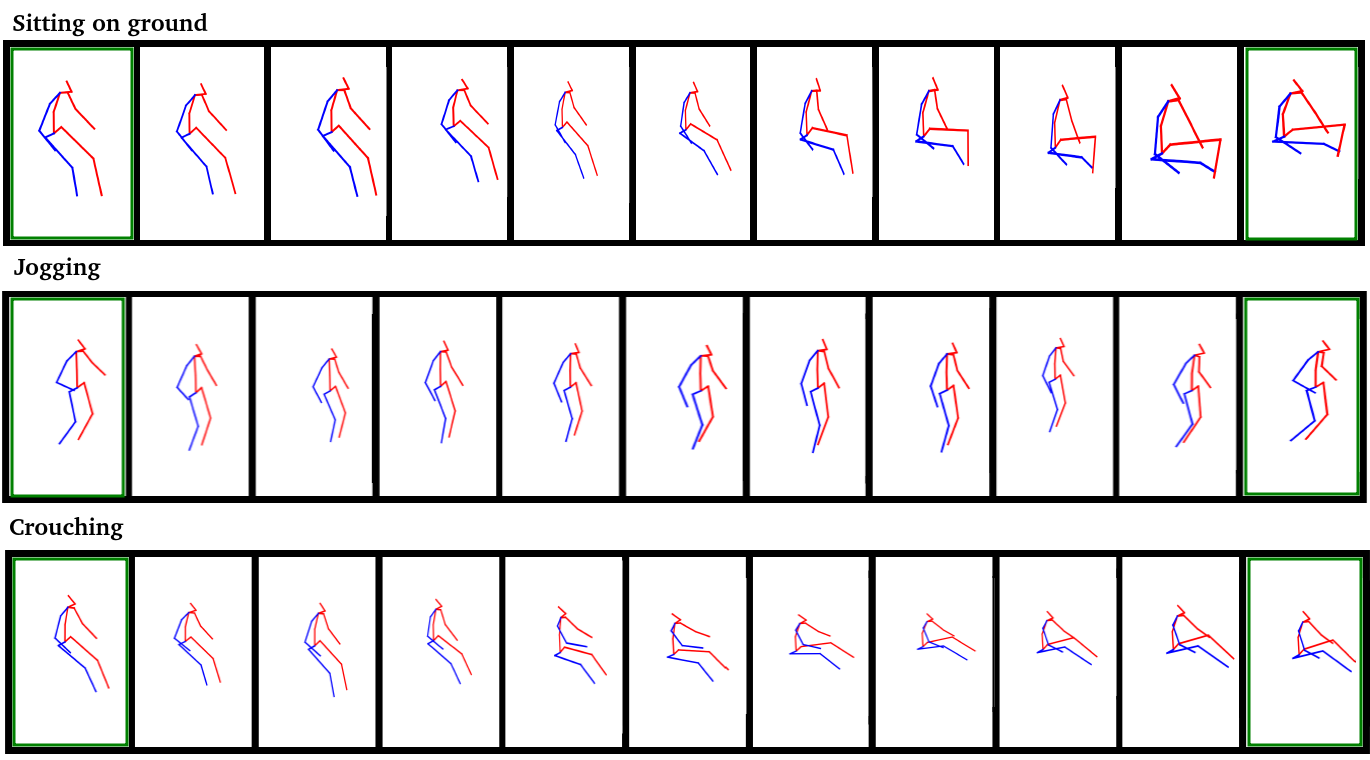}
\end{center}

\caption{Examples of transversal for the activities \textit{sit to ground}, \textit{jogging}, and \textit{crouching}. The endpoints are highlighted in green. }
\label{fig:transversals}
\end{figure*}

Finally, to get insights on the smoothness of the shared representation space, we analyzed several straight Euclidean transversals.
As endpoints, we considered two frames of the same video, $x_i$ and $x_j$ and computed their projection onto the joint space, say $z_i$ and $z_j$. 
We then obtained the corresponding skeletons $p_i$ and $p_j$ by using a supervised 3D egopose method (DeconvNet) pre-trained on the same dataset (\textit{First2Third-Pose}).
Afterwards, we obtained points on the same line in the joint embedded space by interpolating between the two endopoints’s corresponding latent vectors as $z_t = z_i + (z_j-z_i)\beta$, where $\beta$ is a real number corresponding to the slope of the line.
The input features for the 3D egopose supervised model, say $\Phi(x_t)$, are also obtained as interpolation from $\Phi(x_i)$ and $\Phi(x_j)$, and fed to the network together with the latent features $z_t$. In Fig.\ref{fig:transversals}, we present some illustrations. The first and last skeletons, highlighted in green, represent the endpoints. The visualizations of the corresponding skeletons lying on such Euclidean transversal, obtained by incrementing $\beta$ from zero to one with step $0.1$,  clearly show that the learned latent space can be considered to a large extent smooth.

\section{Conclusion}
\label{sec:conclusion}
In this paper, we explored for the first-time how to exploit the link between first- and third-view perspective for the task of egocentric 3D pose estimation.
We proposed a versatile framework to build image features that help to discriminate different 3D human poses from egocentric videos even in a target dataset different than the source dataset used to obtain the joint embedded space. Additionally, we built and made publicly available  \textit{First2Third-Pose}, a large and synchronized dataset of first- and third-view videos capturing 14 people performing overall 40 different activities. Currently, this is the only 3D pose dataset with synchronized first and third-views videos.
To bridge the heterogeneity gap between the two views, we proposed a self-supervised representation learning approach that learns to transform data samples from different views into a common embedding space, which is subsequently employed to extract features from unpaired and unseen egocentric videos. 
We provided insights into the structure of the joint learned feature space, through both data visualization and data analytical tools. These insights suggest that the learned feature space well separate different actions and may be therefore potentially useful also for skeleton based action recognition.
We tested our approach on three state-of-the-art methods and two real datasets. Experimental results demonstrated that the joint embedding space learned with \textit{First2Third-Pose} can be used to enhance supervised state-of-the-art egopose estimation methods on different datasets, without the need of domain adaptation nor knowledge of camera parameters. 
Further research will investigate how to further close the gap between first- and third-view, and how to benefit both first- and third- view domains.

\section{Acknowledgements}
This work has been partially supported by projects PID2020-120049RB-I00 and PID2019-110977GA-I00 funded by MCIN/ AEI /10.13039/501100011033 and by the "European Union NextGenerationEU/PRTR", as well as by grant RYC-2017-22563 funded by MCIN/ AEI /10.13039/501100011033 and by "ESF Investing in your future", and network RED2018-
102511-T funded by MCIN/ AEI.

\bibliography{top}
\end{document}